\documentclass{article}

\usepackage[cloudcomputing]{cosparws_abs} 

\usepackage{longtable}

\usepackage{subcaption}
\usepackage{graphicx}
\usepackage[utf8]{inputenc} 
\usepackage[T1]{fontenc}    
\usepackage{hyperref}       
\usepackage{url}            
\usepackage{booktabs}       
\usepackage{amsfonts}       
\usepackage{nicefrac}       
\usepackage{microtype}      

\usepackage{
  amsmath,
  amssymb,
  amsthm,
  array,
  asymptote,
  enumerate,
  enumitem,
  fancyhdr,
  geometry,
  graphicx,
  listings,
  lstautogobble,
  multicol,
  multirow,
  scrextend,
  tcolorbox,
  xcolor,
  xpatch,
  xspace
}

\title{CELESTIAL: \\ Classification Enabled via Labelless Embeddings with Self-supervised Telescope Image Analysis Learning}

\usepackage{xcolor}

\author{%
  Suhas Kotha \thanks{Work done as a researcher at SpaceML} \\
  Carnegie Mellon University \\
  \texttt{suhask@andrew.cmu.edu} \\
  \AND
  Anirudh Koul \\
  \And
  Siddha Ganju \\
  \And
  Meher Kasam
}
\begin{document}

\maketitle

\begin{abstract}
A common class of problems in remote sensing is scene classification, a fundamentally important task for natural hazards identification, geographic image retrieval, and environment monitoring. Recent developments in this field rely label-dependent supervised learning techniques which is antithetical to the 35 petabytes of unlabelled satellite imagery in NASA GIBS. To solve this problem, we establish CELESTIAL–a self-supervised learning pipeline for effectively leveraging sparsely-labeled satellite imagery. This pipeline successfully adapts SimCLR, an algorithm that first learns image representations on unlabelled data and then fine-tunes this knowledge on the provided labels. Our results show CELESTIAL requires only a third of the labels that the supervised method needs to attain the same accuracy on an experimental dataset. The first unsupervised tier can enable applications such as reverse image search for NASA Worldview (i.e. searching similar atmospheric phenomenon over years of unlabelled data with minimal samples) and the second supervised tier can lower the necessity of expensive data annotation significantly. In the future, we hope we can generalize the CELESTIAL pipeline to other data types, algorithms, and applications.
\end{abstract}

\section{Introduction}

The task of scene classification on satellite imagery is an incredibly fundamental task to many fields. Better insights to this task lend themselves to solutions for preemptively identifying natural hazards, retrieving images of a certain geography, and monitoring ecologies. More importantly, scene classification acts as a huge barrier to scientific inquiry as the lack of automated solutions for identifying relevant data force researchers to spend valuable time sifting through data by hand; this is the converse of the data lending itself to the answer.

As such, there has been a line of work on algorithmic techniques to leverage insight from massive and varied datasets. Of these, the most successful in recent history has been machine learning and deep learning techniques \cite{MA2019166}. In the last decade, deep learning has proved to be incredibly effective at leveraging and generalizing insights from labelled data sets to previously unseen examples. However, most traditional deep learning techniques rely on labelled data; without having a class label associated with each piece of data, the algorithm has a difficult time making sense of this algorithm \cite{sun2017revisiting}.

Unfortunately, this makes it difficult to apply traditional deep learning techniques in scene classification. In this particular domain, it is very easy to accumulate satellite images of the earth in a number of varying types and resolutions. However, the process of labelling scenes is quite expensive. This tension is best seen through the fact that the NASA Global Imagery Browsing Service has 35 petabytes of unlabelled satellite data and is projected to have 250 petabytes by 2050. It took ImageNet 22 human-years to label their 14 million images, and GIBS is already $200,000$ times bigger. Surely, labelling a sizeable portion of this data set is quite expensive and doesn't leverage the full information gain of the archive. As such, we are focused on methods to leverage insights from this data set. Our approach utilizes self-supervised learning to properly utilize sparsely labelled data. Our key contributions can be summarized as follows.
\begin{enumerate}
    \item An unsupervised image featurizer that can effectively retrieve similar looking images without any prior label knowledge
    \item A methodology that enables few-shot image training of downstream tasks that are historically label-reliant
    \item A proposal for a framework to empower scientists in automatically generating rare phenomena data sets for their research projects
\end{enumerate}

\section{Preliminaries}

\subsection{Supervised Learning}

The supervised learning framework is the traditional formulation under which machine learning problems are formulated. In this paradigm, an algorithm gets access to a data set where each sample has an associated label. The algorithm heavily relies on this label to make sense of the given sample.

For example, we can consider the Empirical Risk Minimization formulation for a supervised parametric model. The algorithm is given $\{(x_i, y_i)\}_{i\in[N]}$, where for $N$ samples, each $x_i$ is the data and $y_i$ is its associated label (this label can be a real number in regression or an element of a set in classification). The model is a function $h$, which, given a set of weights $W$, attempts to map an input sample to its associated label. We also define a loss function $\ell$, or a way to measure the quality of an estimate for a label. Once these are defined, the objective of a supervised algorithm is 
$$\min_{W} \frac{1}{N}\sum_{i\in[N]} \ell\left(h(x_i, W), y_i\right)$$
With an appropriate choice of $\ell$ (and accounting for the regularization of weights), this description represents the objective of algorithms like ridge regression, decision trees, and deep neural networks. However, as apparent in the formulation, it requires a label for each sample, which makes it an extremely restrictive framework.

\subsection{Unsupervised Learning}

The unsupervised learning framework is a formulation which has grown due to the need for algorithms that work where there is absolutely no labels provided for the data. In this paradigm, since the algorithm does not have an understanding of what the labels could possibly even be, the algorithm needs to interpret the samples in relation to other samples. As such, unsupervised algorithms are focused on relational tasks such as clustering and dimensionality reduction.

For example, we can consider the Empirical Risk Minimization formulation for an unsupervised clustering algorithm. The algorithm is given $\{x_i\}_{i\in[N]}$ which represents $N$ samples. A partition of the data set is an assignment of exactly one label to each sample. We also define a risk function $R$ that can assess the quality of a partition. Once these are defined, the objective of an unsupervised algorithm is 
$$\min_{C} R(X, C)$$
With an appropriate choice of $R$, this description represents clustering algorithms such as k-means and hierarchical linkage. However, as apparent in the formulation, it can not use labels even if they are provided, which makes it an extremely loose framework.

\subsection{Self-supervised Learning}

Self-supervised learning is the effective and elegant combination of both tactics. In this paradigm, the algorithm is given a set of samples with labels and a set of samples without labels. An effective self-supervised algorithm's performance would not be heavily damaged by a lower number of labelled samples and would be benefited by a higher number of unlabelled samples.

Most self-supervised algorithms start with an unsupervised task over the entire data set, minimizing the risk over a "game" to help with learning. After this base knowledge is formed, it is fine-tuned to the labelled subset of the data using a traditional supervised learning algorithm. If we choose to represent our predictor as $f$, we can decompose it into $f = w \circ \phi$. The featurizer is $\phi$, trained purely on the unsupervised game. The classifier is $w$, trained purely on the downstream supervised task \cite{gulrajani2020search}. 

In recent years, there has been many successful self-supervised learning algorithms such as Contrast Predictive Coding \cite{oord2018representation}, DeepCluster \cite{caron2018deep}, and Pretext Invariant Representation Learning \cite{misra2020self}. Among these, we find that Simple Contrastive Learning of Visual Representations (SimCLR) is the most well-equipped technique for our domain \cite{chen2020simple}. We detail the algorithm, its results, and its implications in the following sections.

\begin{figure}[htp]
    \centering
    \includegraphics[width=10cm]{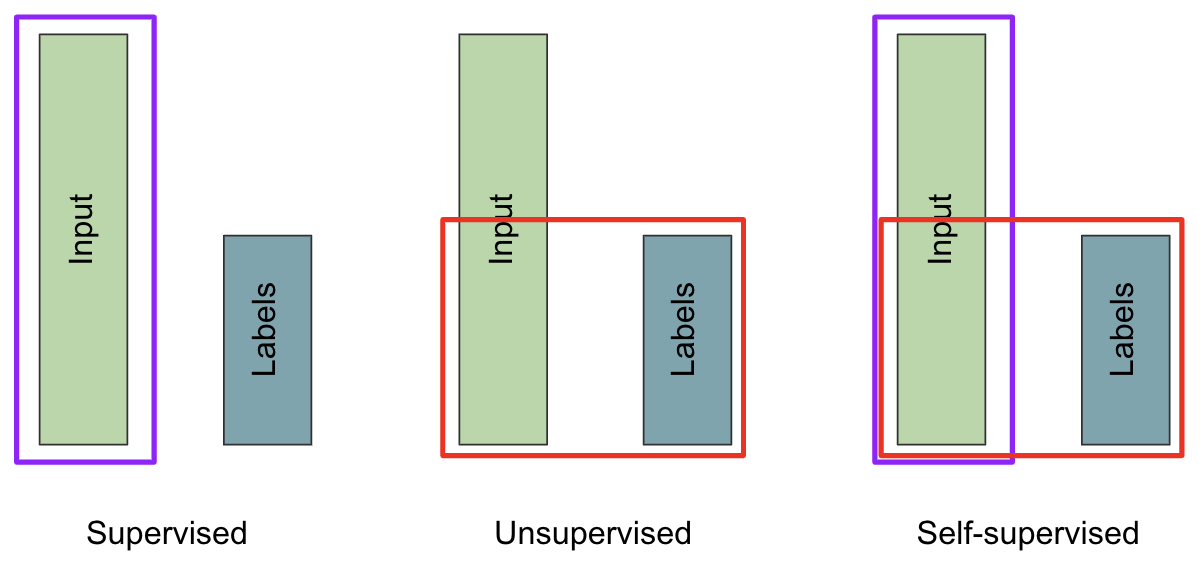}
    \caption{Supervised learning can not use unlabelled data and unsupervised learning can not use labeled data}
    \label{fig:pipeline}
\end{figure}

\section{Pipeline}

\subsection{Tier 1: Unsupervised}

For this portion of the algorithm, we want to train a featurizer over an educational game. The attribute that we try to optimize for over here is \emph{augmentation invariance}. Simply put, we know that small augmentations such as a shift in brightness will be \emph{content-preserving} in that they don't change the class of the image.

\begin{figure}[htp]
    \centering
    \includegraphics[width=12cm]{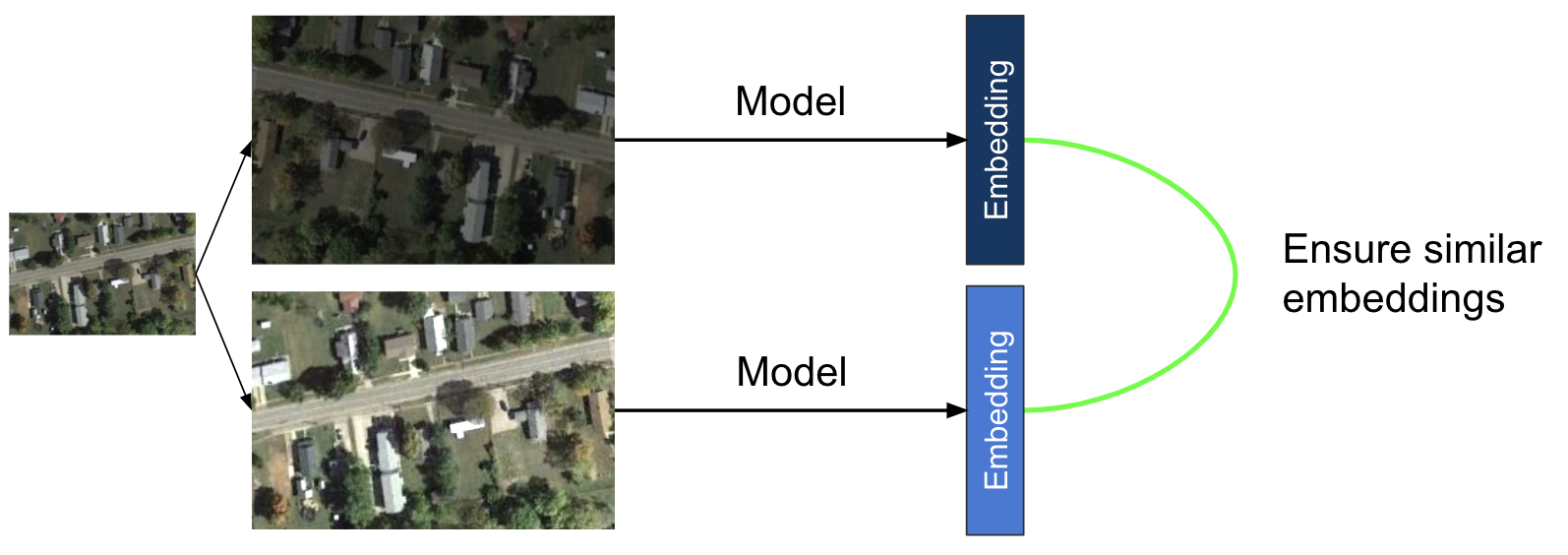}
    \caption{The featurizer learns to embed images such that similar looking images are closer}
    \label{fig:tier1exp}
\end{figure}

We can teach this notion by taking an image and applying two random augmentations on it. The two resulting images should still have the same class. Therefore, we train our featurizer to recognize these images as the same. More formally, if we take an $x$ and generate random perturbations $\tilde{x}^{(1)}$ and $\tilde{x}^{(2)}$, we want to minimize the gap between $\phi(\tilde{x}^{(1)})$ and $\phi(\tilde{x}^{(2)})$ defined via cosine similarity difference. The procedure is outlined in Figure $\ref{fig:tier1exp}$. We trained the network and removed the last few layers, which will be replaced by the supervised head.

\subsection{Tier 2: Supervised}

The supervised portion works quite simply. Given this featurizer, we attach a trainable head to the model. Assuming the featurizer is trained accurately, we can now train the $w$ component of the model by using typical supervised training on the model $w \circ \phi$ (with $\phi$ frozen). The crucial observation is that even though this model has the full expressive power of a deep neural network, it does not have the data to support its inferences. The featurizer, trained for dimensionality reduction, has an implicit sense of clusters within the data. With very few samples, it can calibrate its understanding and provide the correct labels to its semblance of clusters.

\section{Results}

\subsection{Phase 1: Unsupervised}

To test this component of the algorithm, we used the RESISC45 data set, a benchmark for Remote Sensing Image Scene Classification created by Northwestern Polytechnical University. The dataset is comprised of 31,500 images with 45 classes. This is a very complex data set with fine-grained classifications such as circular farmland versus rectangular farmland. The data set was chosen due to its inherent complexity and large sample size, allowing us to experiment on the effects of data volume.

\begin{figure}[htp]
    \centering
    \includegraphics[width=8cm]{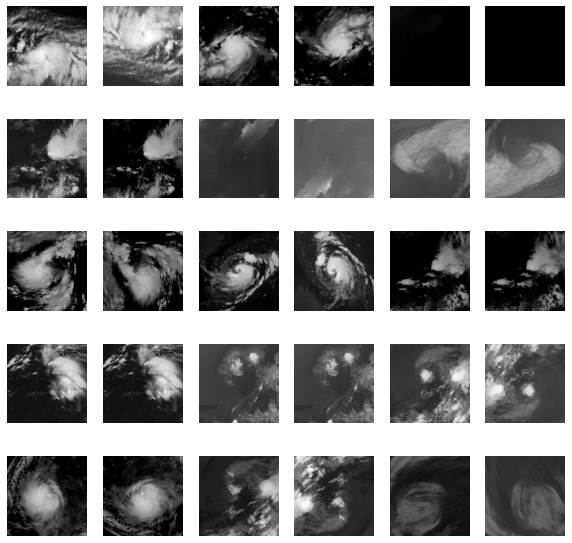}
    \caption{A sample panel where different original images are displayed after a random augmentation}
    \label{fig:augment}
\end{figure}

We used a deep CNN with approximately $500,000$ parameters. For our augmentations, we found that the best combination was random rotation, random brightness change of up to $20\%$, and random contrast change of up to $20\%$. A sample panel of augmentation is displayed in Figure $\ref{fig:augment}$.

\begin{figure}[htp]
    \centering
    \includegraphics[width=12cm]{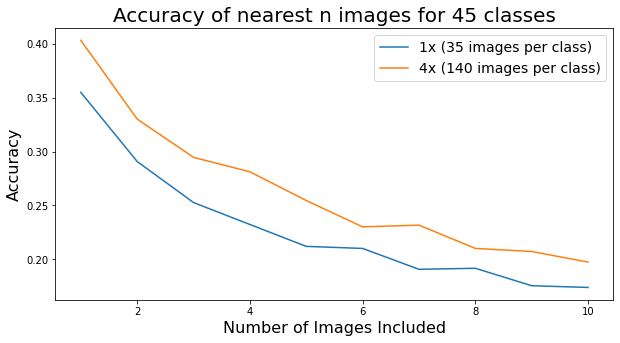}
    \caption{This plot measures the accuracy in the prediction of the $k$th nearest neighbor}
    \label{fig:tier1knn}
\end{figure}

We found that the resulting featurizer was quite performant. Our method of evaluation was through performing a K Nearest Neighbors evaluation purely on the embedding space generated through the game. In Figure $\ref{fig:tier1knn}$, we can see that the model can perform a good job at K Nearest Neighbors even with such a high number of classes. We also see that the performance dramatically increases when given a higher amount of data, proving our featurizer maintains the desired property of improving with a higher volume of unlabelled samples. 

\subsection{Phase 2: Supervised}

To test this component of the algorithm, we used the UC Merced Land Use Data Set. The data set is comprised of 2,100 images with 21 classes. This data set is similarly complicated. The data set was chosen due to its small sample size, allowing us to experiment on the ability of the model to generalize with a small labelled fraction alongisde a small labelled count.

We take the neural network trained using the Phase 1 component and retrain it with the frozen featurizer applied before a trainable head of 2 dense layers. Here, we control what fraction of the data set we give labels for. This allows us to observe the effects of lowered sample on the model's ability to generalize. We benchmark this model against what a purely supervised model would learn from just the labelled data. Though we always expect our algorithm to do better purely since it leverages more information, our goal is to find out by how much. 

\begin{figure}[htp]
    \centering
    \includegraphics[width=12cm]{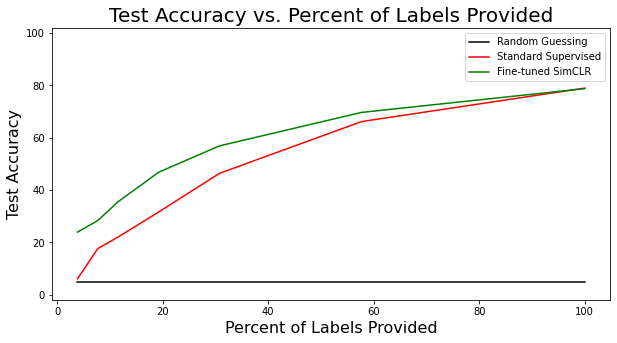}
    \caption{This plot measures the accuracy in scene classification as a function of labels provided}
    \label{fig:tier2knn}
\end{figure}

As you can see in Figure $\ref{fig:tier2knn}$, the model performance does drastically decreases as less labels are provided. However, the self-supervised performance decreases much less rapidly. At the extreme point of $4\%$ of labels, the model performance is impressively $>25\%$, which is much better than the classical training method which is as accurate as random guessing. For the classical model to perform as well as the our model at $4\%$ of labels, it would require three times as many labels. This difference will only be exaggerated as we apply this on larger data sets with larger unlabelled corpora.

\section{Conclusion}

In this work, we have demonstrated the performance and versatility of the CELESTIAL pipeline for self-supervised scene classification. We have also demonstrated the growth potential of this algorithm, as it is scalable with a commodity we will be receiving more of-unlabelled data. This pipeline offers the services of dimensionality reduction, unsupervised nearest neighbor search, and fine-tuning ability for downstream tasks. We can see that self-supervised learning is cheaper, more accurate, and more scalable. 

We believe there are many important applications for this pipeline. This technique should be scalable to multiresolutional and multispectral imagery, allowing it to generalize to a whole class of untapped measurements. Moreover, with recent advances in self-supervised learning, we should be able to improve the first tier in a blackbox manner, furthering performance in the second tier.

Most importantly, we have the vision for an effective image-based search engine. Users, who are presumably investigating a certain phenomenon, could input samples of their phenomena through a web platform. Using this, our intelligent base featurizer can fine-tune for the task at hand and find more samples of what they're looking for. In an interactive manner, the user can use their expertise to approve or decline the model, which only further feeds the model data. This system can accelerate the process of scientific research significantly and can hopefully empower researchers to conduct studies without being limited by their resources. We hope that our work can contribute to this grand vision.

\newpage
\bibliographystyle{plain}
\bibliography{references}

\end{document}